# On the Performance of Metaheuristics: A Different Perspective

Hamid Reza Boveiri [*] and Raouf Khayami

Computer Engineering and Information Technology Department, Shiraz University of Technology
Shiraz, Iran.
[*] E-mail: hr.boveiri@{sutech.ac.ir, gmail.com} — Phone: +98-901-978-0878 — Fax: +98-61362-3811

## *Abstract*

Nowadays, we are immersed in tens of newly-proposed evolutionary and swam-intelligence metaheuristics, which makes it very difficult to choose a proper one to be applied on a specific optimization problem at hand. On the other hand, most of these metaheuristics are nothing but slightly modified variants of the basic metaheuristics. For example, Differential Evolution (DE) or Shuffled Frog Leaping (SFL) are just Genetic Algorithms (GA) with a specialized operator or an extra local search, respectively. Therefore, what comes to the mind is whether the behavior of such newly-proposed metaheuristics can be investigated on the basis of studying the specifications and characteristics of their ancestors. In this paper, a comprehensive evaluation study on some basic metaheuristics i.e. Genetic Algorithm (GA), Particle Swarm Optimization (PSO), Artificial Bee Colony (ABC), Teaching-Learning-Based Optimization (TLBO), and Cuckoo Optimization algorithm (COA) is conducted, which give us a deeper insight into the performance of them so that we will be able to better estimate the performance and applicability of all other variations originated from them. A large number of experiments have been conducted on 20 different combinatorial optimization benchmark functions with different characteristics, and the results reveal to us some fundamental conclusions besides the following ranking order among these metaheuristics, {ABC, PSO, TLBO, GA, COA} i.e. ABC and COA are the best and the worst methods from the performance point of view, respectively. In addition, from the convergence perspective, PSO and ABC have significant better convergence for unimodal and multimodal functions, respectively, while GA and COA have premature convergence to local optima in many cases needing alternative mutation mechanisms to enhance diversification and global search.

## 1. INTRODUCTION

Most of the combinatorial optimization problems in economics, engineering and industry are of such problems cannot be solved using heuristics or exact methods in timely and efficient way because of the NP-hard nature of their search space. Meanwhile, other interesting paradigms inspired by biology and artificial life were introduced named metaheuristics with significant power and potential to cope with such kinds of difficult problems. In such methods, which most of them are originated from natural phenomena, the guarantee of finding exact optimal solutions is sacrificed in favor of obtaining suboptimal solutions in a significantly reduced amount of time. Based on the different criteria being considered, such as population-based, iterative based, stochastic, deterministic, etc., these algorithms can be classified into the different groups. Of them, two important groups introduced in the literature are Evolutionary Algorithms (EA) and Swarm Intelligence (SI).

The most recognized evolutionary algorithm is Genetic Algorithm (GA) introduced by Holland in 1975 (Holland, 1975). GA works on the principle of the Darwinian theory of the survival of the fittest, and the theory of evolution of the living beings. In natural evolution, each species searches for beneficial adaptations in a dynamic environment based on the environmental feedback. As the species evolve, some new attributes are encoded in the chromosomes of their individual members. Although this information does change by random mutation, the real driving force behind the evolutionary development is the combination and exchange of chromosomal material during breeding. Of the other evolutionary algorithms are Differential Evolution (DE) introduced by Storn and Price in 1997 (Storn and Price, 1997), which is similar to GA, but with a specialized crossover and selection method, Evolution Strategy (ES) introduced by Rechenberg in 1973 (Rechenberg, 1973), Evolution Programming (EP) introduced by Fogel *et al.* in 1966 (Fogel *et al.*, 1966), Artificial Immune Algorithm (AIA) introduced by Farmer *et al.* in 1986 (Farmer *et al.*, 1986) which works on the basis of immune system of the human being, Shuffled Frog Leaping (SFL) introduced by Eusuff and Lansey in 2003 (Eusuff and Lansey, 2003), which works on the principle of communication among the frogs, and Bacteria Foraging Optimization (BFO) introduced by Passino in 2002 (Passino, 2002), which works on the behavior of bacteria.

On the other hand, some well-known swarm intelligence based algorithms are Particle Swarm Optimization (PSO) introduced by Kennedy and Eberhart in 1995 (Kennedy and Eberhart, 1995), which works on the foraging behavior of the swarm of birds, Ant Colony Optimization (ACO) introduced by Dorigo *et al.* in 1991 (Dorigo *et al.*, 1991), which works on the foraging behavior of the real ant for food, and Artificial Bee Colony (ABC) algorithms introduced by Karaboga and Dervis in 2005 (Karaboga and Dervis, 2005) which works on the foraging behavior of a honey bee, Cuckoo Optimization Algorithm (COA) first introduced by Rajabioun in 2011 (Rajabioun, 2011), inspired from the exotic lifestyle in parasite-brooding of a bird family called cuckoo. Besides the evolutionary and swarm intelligence based algorithms, there are some other algorithms which work on the principles of different natural phenomena. Some of them are the Harmony Search (HS) algorithm introduced by Geem *et al.* in 2001 (Geem *et al.*, 2001), which works on the principle of music improvisation in a music player, the Gravitational Search Algorithm (GSA) introduced by Rashedi *et al.* in 2009 (Rashedi *et al.*, 2009), which works on the principle of gravitational force acting between the bodies, Biogeography-Based Optimization (BBO) introduced by Simon in 2008 (Simon, 2008), which works on the principle of immigration and emigration of the species from one place to the other, the Grenade Explosion Method (GEM) introduced by Ahrari and Atai in 2010 (Ahrari and Atai, 2010), working based on the principle of explosion of a grenade, the League Championship Algorithm (LCA) introduced by Kashan in 2011 (Kashan, 2011), the Charged System Search (CSS) introduced by Kaveh and Talatahari in 2010 (Kaveh and Talatahari, 2010), and Teaching-Learning-Based Optimization (TLBO) algorithm introduced by Rao *et al.* in 2011 (Rao *et al.*, 2011), working based on the interacting behavior of a teacher and some learners in a classroom.

It is clear that some of the aforementioned metaheuristics are basic and underlying algorithms to inspire and propose the others in the literature. We should be aware of that a derived algorithm such as DE or SFL, which are originated from GA, inherit most of their properties from their parent. To exemplify, in (Adam *et al*., 2017), it has been proved that most of the DE variants are efficient only if there is a large initial population (sometimes up to 1000 individuals!) as well as large number of iterations and Fitness Function Evaluations (FFEs); this property is also valid for the parent of DE i.e. GA to a large extent. Of course, we have to be familiar with these properties to select a proper metaheuristic for applying on the specific optimization problem at hand. On this basis, a comprehensive comparison and evaluation study on these basic metaheuristics will open up a new insight into a deeper estimation of the performance and applicability of a large number of newly proposed variations of such metaheuristics. In this paper, a comprehensive evaluation study on some basic metaheuristics i.e. GA, PSO, ABC, TLBO and COA is conducted, and the results is analyzed from two different perspectives i.e. performance and convergence speed. Actually, introducing and comparing newly-proposed metaheuristics, their structures and performance are out-of-agenda for this paper, and we suggest referring to (Adam *et al*., 2017) and (Nabaei *et al*., 2016) or follow CEC'2017 optimization competitions (IEEE CEC, 2017) for this purpose. Instead, the readers of this study can find the answers to the following questions by reading this paper: 1) what is the overall performance ranking among the basic metaheuristics? 2) which metaheuristic or its extended variations is suggested to be applied on simpler engineering optimization problems with unimodal search space? 3) which metaheuristic or its extended variations is suggested to be applied on more complex optimization problems with multimodal search space? 4) what are the convergence natures of the basic metaheuristics? 5) which metaheuristic or its extended variations is suggested to be applied where the time is not convenience for the problem at hand such as real-time applications/communications?

The rest of the paper is organized as follows. The basic metaheuristics under the consideration in this study are described in the following Section. Section 3 explains the implementation details and configurations. Section 4 is devoted to the achieved experimental results and comparison study, and finally the paper is concluded in the last Section.

## 2. Metahuristic Algorithms to Be Studied and Evaluated

This Section is devoted to introduce those conventional basic metaheuristics among which the comparison study will be made i.e. GA, PSO, ABC, TLBO and COA. For a rational judgment, either an evolutionary computation algorithm i.e. GA or four swarm intelligence-based algorithms named PSO, ABC, TLBO and COA are considered and described as follows:

### 2.1. Genetic Algorithm (GA)

GA was first introduced by Holland in 1975 (Holland, 1975). It is a stochastic searching method based on the theory of Darwinian natural evolution of the living beings. This algorithm is started with a set of randomly generated solutions called initial population. Each member in the population is called a chromosome which is actually a solution of the given problem, and itself consists of a string of genes. The number of genes in each chromosome, and their acceptable value's ranges are depended on the problem specification e.g. in combinatorial function optimization, the number of genes for each chromosome is corresponding to the number of optimization variables, and the gene values are bounded by an upper and lower bounds of these variables. A set of chromosomes (population) in each iteration is called a generation. The generation is evaluated by a fitness function in order to find the desirability of each individual. Afterwards, some offspring (the new generation) is created by applying some operators on the current generation. These operators are crossover which selects two chromosomes as parents, combines them and generates two new offspring, and mutation which changes randomly value of some genes in a selected chromosome, and creates a new offspring (Fig. 1). Then, the best children and maybe their parents are selected by evolutionary selection operator according to their fitness values using methods like ranking, roulette-wheel, tournament and so on. These three phases of production, i.e. manipulation, evaluation and selection, are repeated until some conditions are satisfied, and finally, the best chromosome in the last generation is returned as the best global solution.

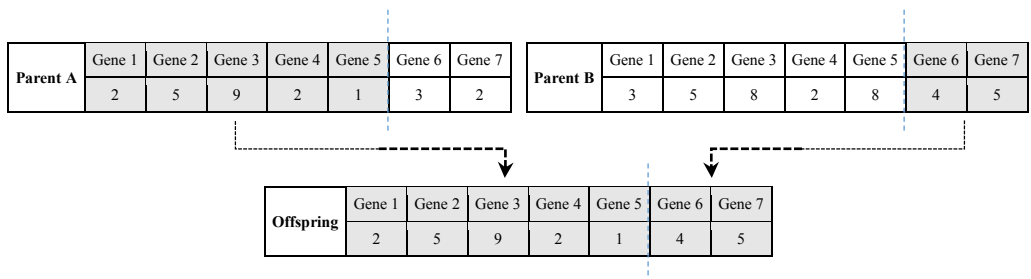

Fig. 1: An example of a single point crossover operation.

### 2.2. Particle Swarm Optimization (PSO)

PSO was first introduced by Kennedy and Eberhard in 1995, originated from the mystery of migration and the foraging behavior of the flocks of birds (called particles) for food (Kennedy and Eberhard, 1995). In this technique, all the particles search for the food in multidimensional search space based on their two important characteristics i.e. the current position referred to as the suggested solution ($x_{i.j}(t)$) and velocity or changing rate of the particle position ($v_{i.j}(t)$) using (1) and (2).

$$x_{i.j}(t) = x_{i.j}(t-1) + v_{i.j}(t) \mid i = 1, 2, \ldots, N_{pop} \text{ and } j = 1, 2, \ldots, N_{var} \tag{1}$$

$$v_{i.j}(t) = w \times v_{i.j}(t-1) + \varphi_1 \left(x_{i.j}^p - x_{i.j}(t-1)\right) + \varphi_2 \left(x_j^g - x_{i.j}(t-1)\right) \tag{2}$$

where $w$ is an inertia factor to tune the velocity in each iteration, $x_i^p$ is the personal best position visited yet by the particle $x_i$, $x^g$ is the global best particle in the population, $\varphi_1 = c_1 \times rand(0.1)$ and $\varphi_2 = c_2 \times rand(0.1)$ are randomly generated personal and global coefficients, respectively, for knowledge exploitation in the algorithm (where $c_1$ and $c_2$ are set to 2 in most the cases). Obviously, if any particle finds a better path to the food's location, it becomes the global best, and attracts other particles to follow its path (global search). On the other hand, each particle exploits its own personal best location as a local search around itself. All particles move slowly towards the obtained solution updating their personal best and the global best solution. At the end, all particles reach the same position supposed to be the best global solution of the given problem.

## 2.3. Artificial Bee Colony (ABC)

ABC algorithm was first introduced by Karaboga and Dervis in 2005, working based on the foraging behavior of a colony of honey bees (Karaboga and Dervis, 2005). In the ABC algorithm, the colony of artificial bees is divided to three groups of different bees just like their real-world counterparts i.e. employed bees, onlookers and scouts. A bee waiting on the dance area for making decision to choose a food source is called an onlooker, and a bee going to the food source previously visited by itself is named an employed bee. On the other hand, a bee carrying out random search to find probably undiscovered food source yet is called a scout. In the ABC algorithm, the first half of the colony consists of employed artificial bees, and the second half constitutes the onlookers. For every food source, there is only one employed bee i.e. each individual employed bee is associated to a certain food source. The employed bee whose food source is exhausted becomes a scout starting new randomly flights around the hive. At the initialization stage, a set of food source positions are randomly selected by the employed bees, and their nectar amounts are determined using the given fitness function. Then, the iterative searching algorithm begins, each cycle of which consists of three steps: 1) each employed bee e.g. $i$-th selects another food source e.g. $k$-th, randomly, and goes toward it from its associated food source using (3).

$$v_{i.j}(t) = x_{i.j}(t) + w_1 rand(-1.1)(x_{i.j}(t) - (x_{k.j}(t)) \mid i = 1, 2, \ldots, N_{pop} \tag{3}$$

where $j \in \{1, 2, \ldots, N_{var}\}$ is a randomly selected index as an individual dimension, and $w_1$ is the migration coefficient used to tune and control global search in ABC. Then, the nectar amount of this location ($v_{i.j}(t)$) is measured; if the fitness value of this location is better than the previous food source location ($x_{i.j}(t)$), it is replaced with newly discovered location; else, it will be remained unchanged. 2) each onlooker bee selects a food source e.g. $i$-th using a roulette-wheel selection based on the nectar amount of the foods. It selects another food source e.g $k$-th randomly, and goes from the $i$-th food source to the selected $k$-th one using (3) again where $w_2$ is used instead. Actually, $w_2$ is a coefficient used to tune and control local search in ABC. At this time, the onlooker bee measures the nectar amounts of the neighborhood ($v_{i.j}(t)$); if the fitness value of the neighborhood is better than the current food source location ($x_{i.j}(t)$), it is replaced with its neighborhood location; else, it will be remained unchanged. 3) determining each scout bee e.g. $i$-th one (each food source that has not been changed for a limited number of iterations simply known as *limit*), and then sending it to search for the potential yet undiscovered food sources using (4).

$$x_{i.j}(t) = x_{min} + rand(0.1)(x_{max} - x_{min}) \mid \text{for every } j = 1, 2, \ldots, N_{var} \tag{4}$$

By means of these simple steps, the bees will converge to the most profitable locations in terms of the given fitness function.

## 2.4. Teaching-Learning-Based Optimization (TLBO)

TLBO algorithm was first introduced by Rao *et al.* in 2011, working based on the interacting behavior of a teacher and some learners in a classroom (Rao *et al.*, 2011). Teaching–learning is an important motivating process where any individual tries to learn something from the others. Traditional classroom teaching–learning environment is one sort of motivating process where the students try to learn from a teacher as well as to share their learned subjects to improve their knowledge. Based on this interacting process, TLBO has been proposed which simulates the traditional teaching–learning phenomenon in a classroom. Actually, TLBO is a population-based algorithm where a group of students (i.e. solution-vectors) is considered, and the different subjects offered to the learners is analogous with the manipulation of different decision variables of the given optimization problem. The algorithm simulates two fundamental modes of learning: 1) learning through teacher known as the teacher phase (global search) and 2) interacting with the other learners known as the learner phase (local search). In each iteration, the best solution in the entire population is considered as the teacher to perform teacher phase, and then leaners starts to share their knowledge to each other as to perform learner phase. In this way, whole the population converge to a same position supposed to be the best global solution of the problem under the consideration.

## 2.5. Cuckoo Optimization Algorithm (COA)

Cuckoo Optimization Algorithm (COA) is a new swarm-intelligence-based metaheuristic first introduced by Rajabioun in 2011 (Rajabioun, 2011), inspired from the exotic lifestyle of a bird family called cuckoo. Specific egg-laying and breeding characteristics of cuckoos called parasite-brooding is the basis of constituting this novel optimization algorithm. Each solution vector in the COA is represented by a "habitat vector" e.g. $H_i = [x_{i,1}, x_{i,2}, …, x_{i,Nvar}]^T$, which is the current location of either a mature cuckoo in the society or an individual egg. Mature cuckoos lay their eggs in some other bird's nests by mimicking their egg's color, pattern and size. In the nature, each cuckoo lays from 5 to 20 eggs, which are suitable as the upper and lower bounds of egg dedication to each cuckoo for most of the problems. Another habit of real cuckoos is that they lay eggs within a limited distance from their habitats called Egg Laying Radius (ELR) (Fig. 2). In an optimization problem with upper and lower bound as $x_{max}$ and $x_{min}$ for the decision variables, respectively, each cuckoo has an ELR which is proportional to the total number of eggs laid by all the cuckoos, the number of current cuckoo's eggs, and also the difference between $x_{max}$ and $x_{min}$. On this basis, the ELR for each cuckoo to lay its eggs is defined as

$$ELR_i = \alpha \times \frac{Negg_i}{Total\ number\ of\ eggs} \times (x_{max} - x_{min}), \quad (5)$$

where $Negg_i$ is the number of eggs laid by the *i*-th cuckoo, and *α* is the egg-laying coefficient, a constant real value supposed to handle the maximum value of ELR.

If these eggs cannot be recognized and killed by the host birds, they grow and become a mature cuckoo, too. Environmental features and the immigration of societies (groups) of cuckoos hopefully lead them to converge and find the best environments for breeding and reproduction using the following migration equation:

$$H_{i.j}^{new} = H_{i.j}^{old} + F \times rand(0.1) \times (H_{best.j} - H_{i.j}^{old}) \mid i = 1, 2, …, N_{pop}\ and\ j = 1, 2, …, N_{var} \quad (6)$$

where *F* is the migration coefficient, and *rand* (0, 1) generates random number in the range of [0, 1]. As it is shown in Fig. 3, it is worth to mention that the cuckoos do not fly all the way to the destination habitat when they move toward the goal point, but they only fly a part of the way (*λ*) and also with some deviation (*φ*). A *λ* randomly selected in range of (0, 1], and *φ* generated randomly in range of (-π/6, + π/6) are introduced suitable for good convergence in the basic COA (Rajabioun, 2011); therefore, the migration coefficient (*F*) should be tuned up according to these bases.

Actually, these the most profitable environments that cuckoos are considered to migrate supposed to be the global optima of the given objective function, and in this way they are able to find optimal solutions for the given optimization problem. The COA has gained an increasing popularity in the past few years, and been applied on the verity of industrial and engineering applications such as task scheduling (Akbari and Rashidi, 2016), cancer classification (Elyasigomari *et.al*, 2017), Prediction of blast-induced ground vibration (Faradonbeh and Monjezi, 2017), and load balancing in transshipment terminal (Bazgosha, Ranjbar and Jamili, 2017).

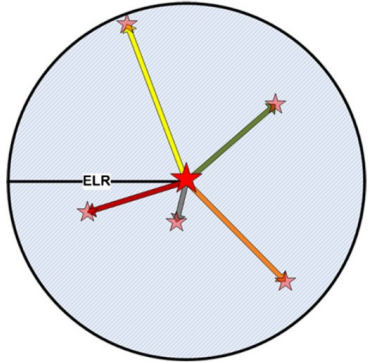

Fig. 2: Random egg laying in ELR, central red star is the initial habitat of the cuckoo with 5 eggs; pink stars are the eggs in new nests (Rajabioun, 2011).

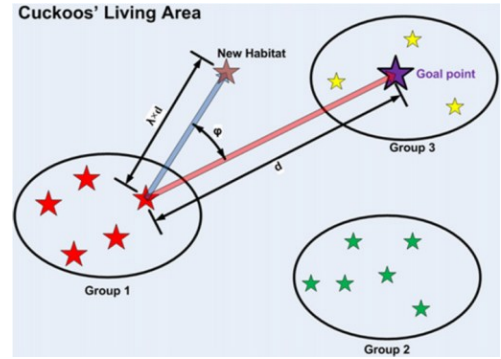

Fig. 3: The immigration of cuckoos in groups toward the globally best habitat (Rajabioun, 2011).

## 3. IMPLEMENTATION DETAILS AND CONFIGURATIONS

The evaluating system to compare the metaheuristics in consideration was implemented on a Pentium IV (8-core 3.9 GHz i7–3770K processor) desktop computer with Microsoft Windows 7 (X64) platform using Microsoft Visual Basic 6.0 programming language. A set of 20 numerical unimodal and multimodal benchmark functions with various search-space structures are considered to fully evaluate the aforementioned approaches. All of these benchmark functions are minimization ones described in the following sub-Section. A full list of configurations used to tune these algorithms is shown in the Table I. Since the overall performance of each algorithm is fully dependent to its configuration, we pay a full attention to this issue, and propose the most adequate configuration for each individual algorithm based on the large number of experiments, and this is considered to be another contribution of this paper. On the other hand, some parameters are identical for all of the utilized algorithms e.g. the population size is set to 40 for all of them (except for COA that is set to 20), the number of iterations is set to 1000, and each algorithm will be terminated after $1333 \times D$ times of Fitness Function Evaluation (FFE), where $D$ is the number of dimensions for the given benchmark function under the experiment. Since in all the experiments in this paper, $D$ is set to 30, the termination criterion is 40,000 FFE for all the algorithms. We believe that, instead of CEC competitions where the maximum number of FFE is often bounded to $10000 \times D$, using $1333 \times D$ times of FFE is fully enough for all the algorithms considered in this study to release their full potential, as it can be seen and proved in Section 4.1 (Convergence Study).

Table I: The different configurations for the different comparison algorithms

| Algorithm | Parameter | Symbol | Value |
|---|---|---|---|
| GA | Mutation Coefficient | $\mu_m$ | 0.9 – each time the mutation is only done on one dimension. |
| | Crossover Coefficient | $\mu_c$ | 0.9 |
| | Selection Mechanism | | Rank-based |
| | Crossover Strategy | | 1-point |
| PSO | Personal Coefficient | $c_1$ | 2.0 |
| | Global Coefficient | $c_2$ | 2.0 |
| | Inertia Factor | $w$ | 0.25 |
| | $rand(-1,1)$ is used instead of $rand(0,1)$ for $\varphi_1$ and $\varphi_2$ in (7). | | |
| ABC | Global Search Coefficient | $w_1$ | 1.0 |
| | Local Search Coefficient | $w_2$ | 1.3 |
| COA | ELR Coefficient | $\alpha$ | 1.0 |
| | Migration Coefficient | $F$ | $\pi/6 \times rand(0, 1)$ |
| | The No. of Clusters | $c$ | 1 |
| | The Minimum No. of Eggs for Each Cuckoo | $Negg_{min}$ | 2 |
| | The Maximum No. of Eggs for Each Cuckoo | $Negg_{max}$ | 5 |
| | Epsilon Tolerance to Kill the Eggs in the Same Locations | $\varepsilon$ | 1.00E-08 |
| | The Population Size | | Half of the others (i.e. 20) |
| TLBO | This algorithm has no parameter to be tuned. | | |

## 3.1. Comparison Benchmark Functions

Table II and Table III list two different sets of 10 multidimensional unimodal and multimodal combinatorial numerical benchmark functions, respectively, which are very popular and applicable in the literature; most of them are adopted in the IEEE Congress on Evolutionary Computation (CEC) competitions of 2013 and 2017 (Liang, 2013; Awad, 2016), for example. By definition, the unimodal functions are those that have only one pick and valley as global optimal point, while the multimodal ones have a number of global and local minima/maxima distributed over the search space, and almost very harder to be solved. Besides, multidimensionality of these functions enable us to conduct the experiments using different number of decision variables ranging from low dimensions (easy-to-be-solved) to the very high ones (hard-to-be-solved).

Table II: Multidimensional Unimodal Combinatorial Benchmark Functions

| No. | Function | Formula | Range | $x^*$ | $F(x^*)$ |
|---|---|---|---|---|---|
| 1 | Sphere | $\min F(x) = \sum_{i=1}^{D} x_i^2$ | [-100, 100] | $[0, 0, …, 0]^T$ | 0 |
| 2 | Rosenbrock | $\min F(x) = \sum_{i=1}^{D-1}[100(x_i^2 - x_{i+1})^2 + (1 - x_i)^2]$ | [-32, 32] | $[1, 1, …, 1]^T$ | 0 |
| 3 | Schwefel N1.2 | $\min F(x) = \sum_{i=1}^{D}\left(\sum_{j=1}^{i} x_j^2\right)^2$ | [-100, 100] | $[0, 0, …, 0]^T$ | 0 |
| 4 | Schwefel N2.21 | $\min F(x) = \max(|x_i|)$ | [-100, 100] | $[0, 0, …, 0]^T$ | 0 |
| 5 | Schwefel N2.22 | $\min F(x) = \sum_{i=1}^{D}|x_i| + \prod_{i=1}^{D}|x_i|$ | [-10, 10] | $[0, 0, …, 0]^T$ | 0 |
| 6 | Step | $\min F(x) = \sum_{i=1}^{D}(|x_i + 0.5|)^2$ | [-100, 100] | $[-0.5, -0.5, …, -0.5]^T$ | 0 |
| 7 | Quartic | $\min F(x) = \sum_{i=1}^{D}(i \times x_i^4) + rand(0.1)$ | [-1.28, 1.28] | $[0, 0, …, 0]^T$ | * |
| 8 | Elliptic | $\min F(x) = \sum_{i=1}^{D}\left(x_i^2 \times (10^6)^{\frac{i-1}{D-1}}\right)$ | [-5.12, 5.12] | $[0, 0, …, 0]^T$ | 0 |
| 9 | BentCigar | $\min F(x) = x_1^2 + 10^6 \times \sum_{i=2}^{D} x_i^2$ | [-5.12, 5.12] | $[0, 0, …, 0]^T$ | 0 |
| 10 | Discus | $\min F(x) = (10^6 \times x_1^2) + \sum_{i=2}^{D} x_i^2$ | [-5.12, 5.12] | $[0, 0, …, 0]^T$ | 0 |

\* The minimum value for the Quartic function is variable based on the generated value by *rand*(0, 1)

Table III: Multidimensional Multimodal Combinatorial Benchmark Functions

| No. | Function | Formula | Range | $x^*$ | $F(x^*)$ |
|---|---|---|---|---|---|
| 11 | **Rastrigin** | $\min F(x) = \sum_{i=1}^{D}[x_i^2 - 10\cos(2\pi x_i) + 10]$ | [-5.12, 5.12] | $[0, 0, ..., 0]^T$ | 0 |
| 12 | **Ackley** | $\min F(x) = e + 20 - 20\exp\left(-0.2\sqrt{\frac{1}{D}\sum_{i=1}^{D}x_i^2}\right) - \exp\left(\frac{1}{D}\sum_{i=1}^{D}\cos(2\pi x_i)\right)$ | [-32, 32] | $[0, 0, ..., 0]^T$ | 0 |
| 13 | **Griewank** | $\min F(x) = \frac{1}{4000}\sum_{i=1}^{D}x_i^2 + \prod_{i=1}^{D}\cos\left(\frac{x_i}{\sqrt{i}}\right) + 1$ | [-600, 600] | $[0, 0, ..., 0]^T$ | 0 |
| 14 | **Schwefel** | $\min F(x) = -\frac{1}{D}\sum_{i=1}^{D}\left(x_i \sin(\sqrt{|x_i|})\right)$ | [-500, 500] | $\pm[\pi(0.5+k)]^2$ | -418.983 |
| 15 | **Weierstrass** | $\min F(x) = \sum_{i=1}^{D}\left(\sum_{j=0}^{20}[0.5^j \cos(2\pi 3^j \times (x_i + 0.5))]\right) - 20 \times \sum_{j=0}^{20}[0.5^j \cos(2\pi 3^j \times 0.5)]$ | [-0.5, 0.5] | $[0, 0, ..., 0]^T$ | 0 |
| 16 | **NCRastrigin** | $\min F(x) = \sum_{i=1}^{D}[y_i^2 - 10\cos(2\pi y_i) + 10] \mid y_i = \begin{cases} x_i & \|x_i\| < 0.5 \\ \frac{round(2x_i)}{2} & \|x_i\| > 0.5 \end{cases}$ | [-5.12, 5.12] | $[0, 0, ..., 0]^T$ | 0 |
| 17 | **Penalized** | $\min F(x) = \frac{\pi}{D}[\sin^2(\pi y_1) + \sum_{i=1}^{D-1}(y_i - 1)^2\{1 + 10\sin^2(\pi y_{i+1})\} + (y_D - 1)^2] + \sum_{i=1}^{D}u(x_i.10.100.4) \mid y_i = 1 + 1/4(x_i + 1). u(x_i. a. k. m) = \begin{cases} k(x_i - a)^m & x_i > a \\ 0 & -a < x_i < a \\ k(-x_i - a)^m & x_i < -a \end{cases}$ | [-50, 50] | $[0, 0, ..., 0]^T$ | 0 |
| 18 | **Penalized2** | $\min F(x) = 0.1[\sin^2(\pi x_1) + \sum_{i=1}^{D-1}(x_i - 1)^2\{1 + \sin^2(3\pi x_{i+1})\} + (x_D - 1)^2 + (1 + \sin^2(2\pi x_D))] + \sum_{i=1}^{D}u(x_i.5.100.4) \mid u(x_i. a. k. m) = \begin{cases} k(x_i - a)^m & x_i > a \\ 0 & -a < x_i < a \\ k(-x_i - a)^m & x_i < -a \end{cases}$ | [-50, 50] | $[0, 0, ..., 0]^T$ | 0 |
| 19 | **Xin-She Yang F4** | $\min F(x) = \left[\sum_{i=2}^{D}\sin^2(x_i) - e^{-\sum_{i=1}^{D}x_i^2}\right] \times e^{-\sum_{i=1}^{D}\sin^2(\sqrt{|x_i|})}$ | [-10, 10] | $[0, 0, ..., 0]^T$ | 0 |
| 20 | **Inverted Vincent** | $\min F(x) = 1 + \frac{1}{D}\sum_{i=1}^{D}\sin(10 \log(|x_i|))$ | [0.25, 10] | $[0, 0, ..., 0]^T$ | 0 |

## 4. THE ACHIEVED RESULTS AND COMPARISON STUDY

Table IV shows the results obtained by each optimization algorithm on the unimodal benchmark functions listed in Table II with a dimension of 30 decision variables. Worth mentioning that each result illustrated in this paper is extracted from 30 independent runs as mean and Standard Deviation (SD) for the algorithm under the consideration. As can be seen, the TLBO outperforms the others by far in this set of experiments. Actually, the rank-sum based raking for the algorithms drown by these experiments is {TLBO, PSO, ABC, GA, and COA}, which indicates that the TLBO is the best, and COA is the worst from the performance point of view.

Table IV: The Results Achieved by the Algorithms on Unimodal Functions (Dimension = 30)

| | | GA | PSO | ABC | TLBO | COA |
|---|---|---|---|---|---|---|
| **Sphere** | Mean: | 3.07E-02 | 1.01E-28 | 7.36E-12 | **1.31E-90** | 8.56E-02 |
| | SD: | 0.008412629 | 2.9698E-28 | 1.63016E-11 | 1.68062E-90 | 0.031097058 |
| | **Rank:** | 4 | 2 | 3 | 1 | 5 |
| **Rosenbrock** | Mean: | 59.20573856 | 35.64582342 | **9.947050402** | 23.90273766 | 50.37957632 |
| | SD: | 28.23859642 | 36.34705511 | 8.50422904 | 0.502877374 | 26.16473546 |
| | **Rank:** | 5 | 3 | 1 | 2 | 4 |
| **Schwefel N1.2** | Mean: | 956.8894976 | 6.48103E-52 | 2.72144E-16 | **8.7636E-174** | 132114.8616 |
| | SD: | 852.71958 | 1.34821E-51 | 8.3935E-16 | 0.0 | 98653.09966 |
| | **Rank:** | 4 | 2 | 3 | 1 | 5 |
| **Schwefel N2.21** | Mean: | 10.60590625 | 4.262234194 | 27.41483971 | **1.17742E-35** | 11.89867221 |
| | SD: | 0.879586902 | 1.084514635 | 4.817762468 | 4.94052E-36 | 4.724000479 |
| | **Rank:** | 3 | 2 | 5 | 1 | 4 |
| **Schwefel N2.22** | Mean: | 1.005158186 | 2.5609E-13 | 1.28889E-05 | **4.855E-44** | 15.58286598 |
| | SD: | 0.183674212 | 4.72262E-13 | 5.44574E-06 | 2.5279E-44 | 7.743056979 |
| | **Rank:** | 4 | 2 | 3 | 1 | 5 |
| **Step** | Mean: | 10.97022622 | **5.88379E-27** | 2.53218E-09 | 4.4781E-08 | 46.80671072 |
| | SD: | 3.501800115 | 1.00916E-26 | 4.63253E-09 | 5.05706E-08 | 16.55761503 |
| | **Rank:** | 4 | 1 | 2 | 3 | 5 |
| **Quartic** | Mean: | 2.17459E-05 | 1.4988E-16 | 3.33067E-17 | **0.0** | 0.000193904 |
| | SD: | 2.08773E-05 | 4.57008E-17 | 2.86658E-17 | 0.0 | 0.000133849 |
| | **Rank:** | 4 | 3 | 2 | 1 | 5 |
| **Elliptic** | Mean: | 1213.614811 | 48.79010576 | 1.55772E-07 | **1.0159E-86** | 36265.6778 |
| | SD: | 672.8818255 | 90.73217672 | 1.72225E-07 | 1.15347E-86 | 14446.76903 |
| | **Rank:** | 4 | 3 | 2 | 1 | 5 |
| **BentCigar** | Mean: | 19893.77169 | 1.76809E-25 | 1.63294E-06 | **2.64285E-84** | 73399.1959 |
| | SD: | 6450.949155 | 3.63978E-25 | 1.26209E-06 | 4.60142E-84 | 40995.21048 |
| | **Rank:** | 4 | 2 | 3 | 1 | 5 |
| **Discus** | Mean: | 189.3995226 | 28.83584 | 4.49287E-08 | **9.54749E-89** | 28.55259352 |
| | SD: | 513.9928242 | 26.06835764 | 8.89152E-08 | 1.35295E-88 | 8.413472576 |
| | **Rank:** | 5 | 4 | 2 | 1 | 3 |
| **Rank-Sum:** | | 41 | 24 | 26 | 13 | 46 |
| **Lexicographic Rank:** | | 4 | 2 | 3 | 1 | 5 |

Nevertheless, as stated in (Li *et al.*, 2013), we should not exclusively rely on these results because most of the benchmark functions have a global minimum in $[0, 0, …, 0]^T$, which can be exploited as a background knowledge for some algorithms to promptly converge to this point. In order to address the issue, Ref. (Liang, Suganthan and Deb, 2005) suggests the utilization of randomly shift-rotated of these benchmark functions. On this basis, another set of experiments were conducted. Table V shows the results obtained by each optimization algorithm on the shift-rotated of previous unimodal functions. Surprisingly, the TLBO not only loses its efficiently versus other methods but also disable to find best solution for any function among other algorithms! The resulting rank-sum based raking for the algorithms drown by this set of experiments is {ABC, PSO = TLBO, GA, and COA}, suggesting the superiority of ABC, and retardation of COA in terms of performance.

Table V: The Results Achieved by the Algorithms on the Randomly Shift-Rotated Unimodal Functions
(Dimension = 30)

| | | GA | PSO | ABC | TLBO | COA |
|---|---|---|---|---|---|---|
| **Sphere** | Mean: | 0.030676378 | **8.18058E-29** | 4.69091E-12 | 1.87272E-07 | 0.094082404 |
| | SD: | 0.008746727 | 1.34589E-28 | 8.11838E-12 | 2.53865E-07 | 0.05317034 |
| | Rank: | 4 | 1 | 2 | 3 | 5 |
| **Rosenbrock** | Mean: | 61.0987845 | 37.57106258 | **9.743094335** | 24.30563695 | 57.2270359 |
| | SD: | 37.84372065 | 38.73174732 | 7.694775458 | 2.25314498 | 27.69644613 |
| | Rank: | 5 | 3 | 1 | 2 | 4 |
| **Schwefel N1.2** | Mean: | 1115.536642 | **1.39473E-48** | 1.9234E-16 | 5.83827E-08 | 139783.022 |
| | SD: | 1009.04402 | 4.22662E-48 | 1.86957E-16 | 1.67974E-07 | 113361.1896 |
| | Rank: | 4 | 1 | 2 | 3 | 5 |
| **Schwefel N2.21** | Mean: | 10.34282207 | **9.693647047** | 30.58646859 | 31.27068723 | 20.79259838 |
| | SD: | 1.875693164 | 15.38797822 | 4.920389627 | 3.744403097 | 4.913544586 |
| | Rank: | 2 | 1 | 4 | 5 | 3 |
| **Schwefel N2.22** | Mean: | 0.996989369 | 4.428137779 | **1.30404E-05** | 0.01200855 | 25.7591043 |
| | SD: | 0.157362998 | 10.10730958 | 4.56787E-06 | 0.021578317 | 7.243019293 |
| | Rank: | 3 | 4 | 1 | 2 | 5 |
| **Step** | Mean: | 10.72734764 | **8.96357E-26** | 1.10539E-09 | 3.15692E-05 | 43.15350058 |
| | SD: | 2.676714481 | 2.75058E-25 | 1.2218E-09 | 3.99374E-05 | 12.53891498 |
| | Rank: | 4 | 1 | 2 | 3 | 5 |
| **Quartic** | Mean: | 7.93032E-06 | 0.017107815 | **3.33067E-17** | 1.1951E-10 | 0.000169865 |
| | SD: | 4.31453E-06 | 0.054099662 | 2.86658E-17 | 1.50557E-10 | 9.12399E-05 |
| | Rank: | 3 | 5 | 1 | 2 | 4 |
| **Elliptic** | Mean: | 1034.359163 | 22641.67664 | **9.64456E-07** | 0.001788973 | 43363.00725 |
| | SD: | 571.428337 | 23389.10844 | 1.42869E-06 | 0.003162159 | 17053.41193 |
| | Rank: | 3 | 4 | 1 | 2 | 5 |
| **BentCigar** | Mean: | 19088.41517 | 6.125539449 | **5.40424E-06** | 1.056550259 | 77836.26624 |
| | SD: | 6849.322476 | 16.4195847 | 8.6117E-06 | 2.592681982 | 21453.33461 |
| | Rank: | 4 | 3 | 1 | 2 | 5 |
| **Discus** | Mean: | 232.723504 | 23.8880058 | **8.62859E-08** | 8.31763E-06 | 34.74057923 |
| | SD: | 334.5629227 | 19.47031341 | 1.79406E-07 | 1.85611E-05 | 10.69517658 |
| | Rank: | 5 | 3 | 1 | 2 | 4 |
| **Rank-Sum:** | | 37 | 26 | 16 | 26 | 45 |
| **Lexicographic Rank:** | | 3 | 2 | 1 | 2 | 4 |

On the other hand, most of the actual engineering optimization problems have multimodal nature, i.e. the objective function under the consideration may have a large number of suboptimal local minima as well as a few identical global ones distributed over the search space. Therefore, it is very important for each algorithm to illustrate a high potentiality to solve such kinds of problems efficiently. Table VI shows the results achieved by each optimization algorithm on the multimodal benchmark functions listed on Table III with a dimension of 30 decision variables. As one can see, the ABC outperforms the others in this set of experiments. It can be say that the rank-sum based raking for the algorithms drown by this set of experiments is {ABC, TLBO, GA, PSO, and COA}, indicating the superiority of ABC versus the others from the performance perspective.

Table VI: The Results Achieved by the Algorithms on Multimodal Functions (Dimension = 30)

| | | GA | PSO | ABC | TLBO | COA |
|---|---|---|---|---|---|---|
| **Rastrigin** | Mean: | 3.999839652 | 73.45610078 | **6.03437E-05** | 14.69520556 | 67.98148885 |
| | SD: | 0.615374489 | 29.76536083 | 0.000190608 | 4.304927659 | 24.31606898 |
| | Rank: | 2 | 5 | 1 | 3 | 4 |
| **Ackley** | Mean: | 1.563654234 | 1.671420002 | 7.4865E-06 | **2.88658E-15** | 3.693481383 |
| | SD: | 0.287783709 | 0.960558013 | 3.97443E-06 | 1.07258E-15 | 0.379272633 |
| | Rank: | 3 | 4 | 2 | 1 | 5 |
| **Griewank** | Mean: | 1.077329269 | 0.034526632 | 0.001762648 | **4.33681E-20** | 1.437366487 |
| | SD: | 0.02081186 | 0.035936495 | 0.005573959 | 2.2857E-20 | 0.111199708 |
| | Rank: | 4 | 3 | 2 | 1 | 5 |
| **Schwefel** | Mean: | 0.00658767 | 3.62591E-17 | 1.42997E-06 | **1.76766E-65** | 4.28115E-17 |
| | SD: | 0.006485682 | 1.013E-16 | 2.01461E-06 | 5.57541E-65 | 1.03681E-16 |
| | Rank: | 5 | 3 | 5 | 1 | 4 |
| **Weierstrass** | Mean: | 2.837354614 | 3.267368397 | **3.63708E-08** | 1.01568E-07 | 19.00749617 |
| | SD: | 0.405983153 | 1.732510766 | 3.22178E-08 | 1.13973E-07 | 3.819742263 |
| | Rank: | 3 | 4 | 1 | 2 | 5 |
| **NCRastrigin** | Mean: | 3.053180478 | 49.77008472 | **0.434927835** | 23.66916675 | 57.69584403 |
| | SD: | 0.782289588 | 15.34500885 | 0.74104605 | 5.764291924 | 18.15991707 |
| | Rank: | 2 | 4 | 1 | 3 | 5 |
| **Penalized** | Mean: | 0.060566235 | 0.249181125 | **2.04594E-10** | 2.7343E-09 | 8.51730668 |
| | SD: | 0.035570121 | 0.380017546 | 1.73979E-10 | 7.34765E-09 | 8.757466939 |
| | Rank: | 3 | 4 | 1 | 2 | 5 |
| **Penalized2** | Mean: | 0.616476232 | 0.247823977 | **1.29719E-09** | 0.459544203 | 54.34260864 |
| | SD: | 0.203488048 | 0.486100008 | 2.50989E-09 | 0.228131294 | 82.29065982 |
| | Rank: | 4 | 2 | 1 | 3 | 5 |
| **Xin-She Yang F4** | Mean: | 1.00 | 1.00 | 1.00 | 1.00 | 1.00 |
| | SD: | 0.0 | 0.0 | 0.0 | 0.0 | 0.0 |
| | Rank: | 1 | 1 | 1 | 1 | 1 |
| **Inverted Vincent** | Mean: | 0.000355299 | 0.139973929 | **1.77965E-05** | 0.034148733 | 0.195707624 |
| | SD: | 5.97435E-05 | 0.000998358 | 2.56955E-05 | 0.000482401 | 0.084655017 |
| | Rank: | 2 | 4 | 1 | 3 | 5 |
| **Rank-Sum:** | | 29 | 34 | 16 | 20 | 44 |
| **Lexicographic Rank:** | | 3 | 4 | 1 | 2 | 5 |

Again, another set of experiments should be conducted using the shift-rotated of aforementioned multimodal benchmark functions. Table VII shows the results obtained by each optimization algorithm in these experiments. For another time, we observe a reordering among the algorithms where the resulting rank-sum based raking for the algorithms drown by this set of experiments is {ABC, GA, PSO, TLBO, and COA}, i.e. ABC has the best performance, while COA is the worst. The results of previous sets of experiments along with this experiments reveal us the following conclusions. Firstly, TLBO has a significant performance on the minimization benchmark functions of which their global optimum is located on $[0, 0, …,0]^T$, while when the global minima is relocated randomly, it lose much of its efficiency! As a consequence, there is a bias movement toward the center of search space in TLBO. We hope this defect had been resolved in the next variations. Secondly, ABC is strongly suggested for complex real-world combinatorial optimization problems since it shows an outstanding performance in both unimodal and multimodal benchmark functions. Thirdly, although PSO is very good on unimodal functions, it gives up its 2nd rank to GA on multimodal benchmark functions; however, PSO is very faster, but GA consume a huge time and resource to be performed so that we do not suggest it where time is not convenient for application e.g. real-time decision or communication. Finally, considering all the experiments on shift-rotated of both unimodal and multimodal benchmark functions leads us to the following ranking, {ABC, PSO, TLBO, GA, COA}.

Table VII: The Results Achieved by the Algorithms on Randomly Shift-Rotated Multimodal Functions (Dimension = 30)

|  |  | GA | PSO | ABC | TLBO | COA |
|---|---|---|---|---|---|---|
| **Rastrigin** | Mean: | 3.083235545 | 77.13585957 | **1.36636E-05** | 44.88763155 | 101.4422262 |
|  | SD: | 0.828532202 | 8.695860847 | 4.30203E-05 | 7.125717498 | 30.86562771 |
|  | Rank: | 2 | 4 | 1 | 3 | 5 |
| **Ackley** | Mean: | 1.708941902 | 1.296393696 | **1.00145E-05** | 7.280337076 | 3.907358314 |
|  | SD: | 0.246079433 | 1.050427098 | 8.10704E-06 | 2.061197124 | 0.61594009 |
|  | Rank: | 3 | 2 | 1 | 5 | 4 |
| **Griewank** | Mean: | 1.096534942 | 0.018676134 | **0.003816886** | 0.064061973 | 1.391340595 |
|  | SD: | 0.027378731 | 0.027685844 | 0.008047465 | 0.090243566 | 0.10741895 |
|  | Rank: | 4 | 2 | 1 | 3 | 5 |
| **Schwefel** | Mean: | 0.005924389 | 5.1418E-16 | 3.14741E-06 | **0.0** | 9.52304E-17 |
|  | SD: | 0.007199123 | 1.58915E-15 | 5.92209E-06 | 0.0 | 2.66053E-16 |
|  | Rank: | 5 | 3 | 4 | 1 | 2 |
| **Weierstrass** | Mean: | 2.82237824 | 7.668301539 | **4.73802E-08** | 10.96305038 | 24.31145159 |
|  | SD: | 0.409797475 | 2.570109955 | 4.122E-08 | 2.564721382 | 3.355842236 |
|  | Rank: | 2 | 3 | 1 | 4 | 5 |
| **NCRastrigin** | Mean: | 2.405475301 | 56.8527291 | **0.300091876** | 46.99844973 | 101.5015375 |
|  | SD: | 0.586761582 | 23.4882275 | 0.482990219 | 9.232907087 | 41.1310273 |
|  | Rank: | 2 | 4 | 1 | 3 | 5 |
| **Penalized** | Mean: | 0.044705363 | 0.405949937 | **3.04326E-10** | 7.272889865 | 20.26999494 |
|  | SD: | 0.038998014 | 0.675565842 | 3.71891E-10 | 6.327039958 | 14.00054352 |
|  | Rank: | 3 | 2 | 1 | 4 | 5 |
| **Penalized2** | Mean: | 0.539863696 | 1.405913819 | **2.72487E-09** | 60.47457633 | 104.0140477 |
|  | SD: | 0.199468796 | 1.5139566 | 6.02866E-09 | 9.111843458 | 138.0667509 |
|  | Rank: | 2 | 3 | 1 | 5 | 5 |
| **Xin-She Yang F4** | Mean: | 1.00 | 1.00 | 1.00 | 1.00 | 1.00 |
|  | SD: | 0.0 | 0.0 | 0.0 | 0.0 | 0.0 |
|  | Rank: | 1 | 1 | 1 | 1 | 1 |
| **Inverted Vincent** | Mean: | 0.000190822 | 0.656290425 | **3.0413E-06** | 0.108911148 | 0.178184464 |
|  | SD: | 4.91952E-05 | 0.036145927 | 4.05053E-06 | 0.105777027 | 0.06615309 |
|  | Rank: | 2 | 5 | 1 | 3 | 4 |
| **Rank-Sum:** |  | 26 | 29 | 13 | 32 | 41 |
| **Lexicographic Rank:** |  | 2 | 3 | 1 | 4 | 5 |

## 4.1. Convergence Study

In order to study the convergence speed of the metaheuristics under the consideration in this study, four unimodal benchmark functions (Sphere, Rosenbrock, BentCigar, and Schwefel N2.22) as well as four multimodal benchmark functions (Ackley, Schwefel, Rastrigin, and Inverted Vincent) among Table II and Table III were considered. The setting and configuration are exactly as same with the previous experiments e.g. the dimension is 30, and the maximum number of FFE is 40000. In case that the algorithms are terminated before the maximum number of FFE is reached is for all the population is trapped in a same point, and no further investigation is possible. Fig. 4 and Fig. 5 are the convergence diagram between the function value (in logarithmic scale) and algorithm iterations (the fitness value of the best solution yet achieved is plotted) for the aforementioned functions. The plotted convergence graphs for both unimodal and multimodal functions reveal to us following conclusions. First of all, GA and COA have terminated before a complete 40000 FFEs in some cases, while others not, and this justifies their occasional degraded results versus PSO, ABC, and TLBO, which is rational based on the no free launch theorem for optimization (Wolpert *et al*., 1997); on this basis, we suggest that they are adequate for those problems needing fast decisions e.g. real-time applications or communications; nevertheless, some extra mutation mechanisms may have a good contribution in their performance, too. Secondly, among PSO, ABC, and TLBO, which exploit all of their accessible FFEs, PSO is very good in unimodal functions while ABC has significant convergences in multimodal ones; hence, we strongly suggest the utilization of PSO in simple engineering problems, where the problem can be modeled as a standard unimodal optimization problem, while applying the ABC on the complex engineering and industrial problems, where the search space has some global optima among a number of local minima/maxima around.

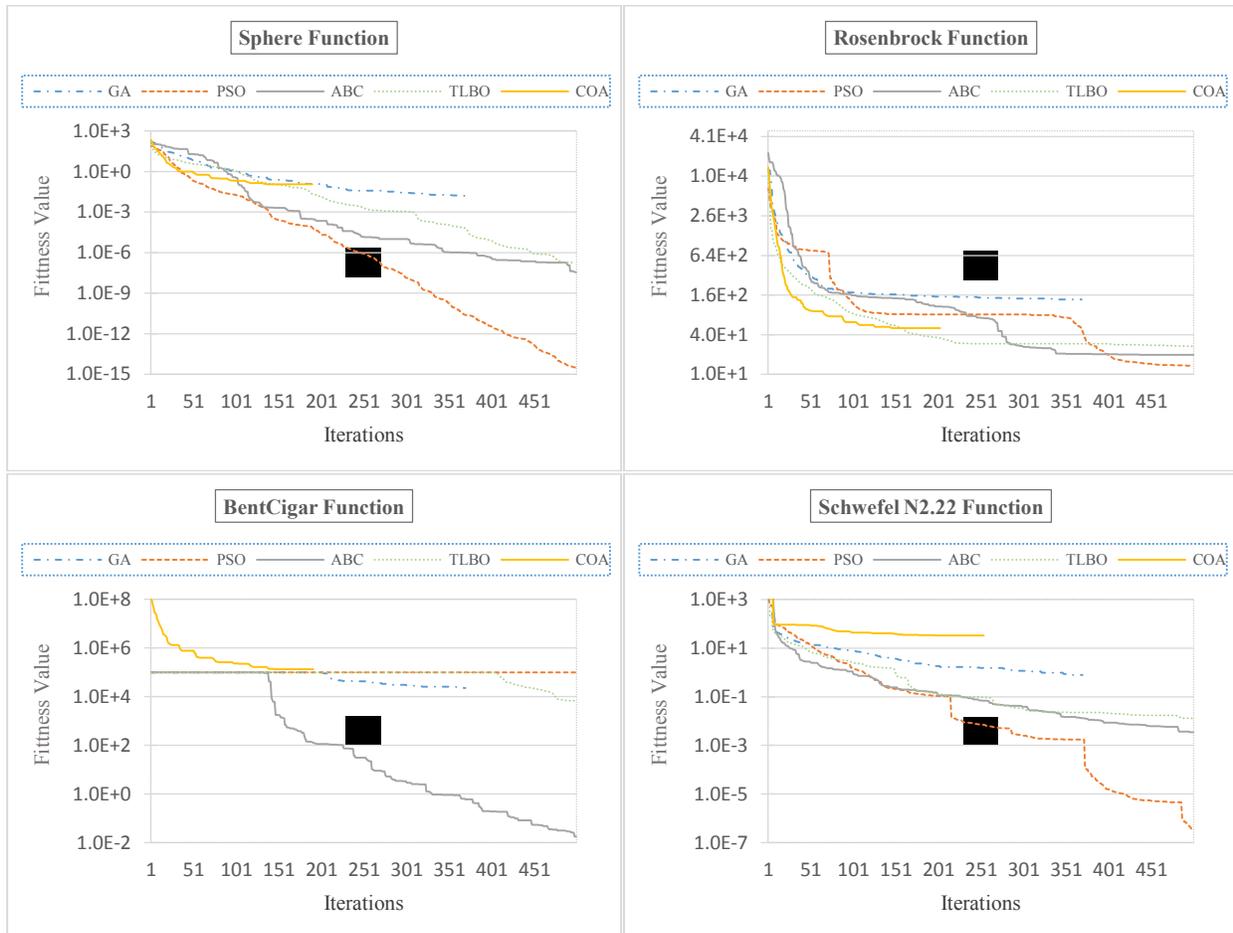

Fig. 4: Convergence diagram of the metahuristics for four unimodal functions with the dimension of 30 (Top-Left: Sphere function, Top-Right: Rosenbrouck function, Bottom-Left: BentCigar function, and Bottom-Right: Schwefel N2.22).

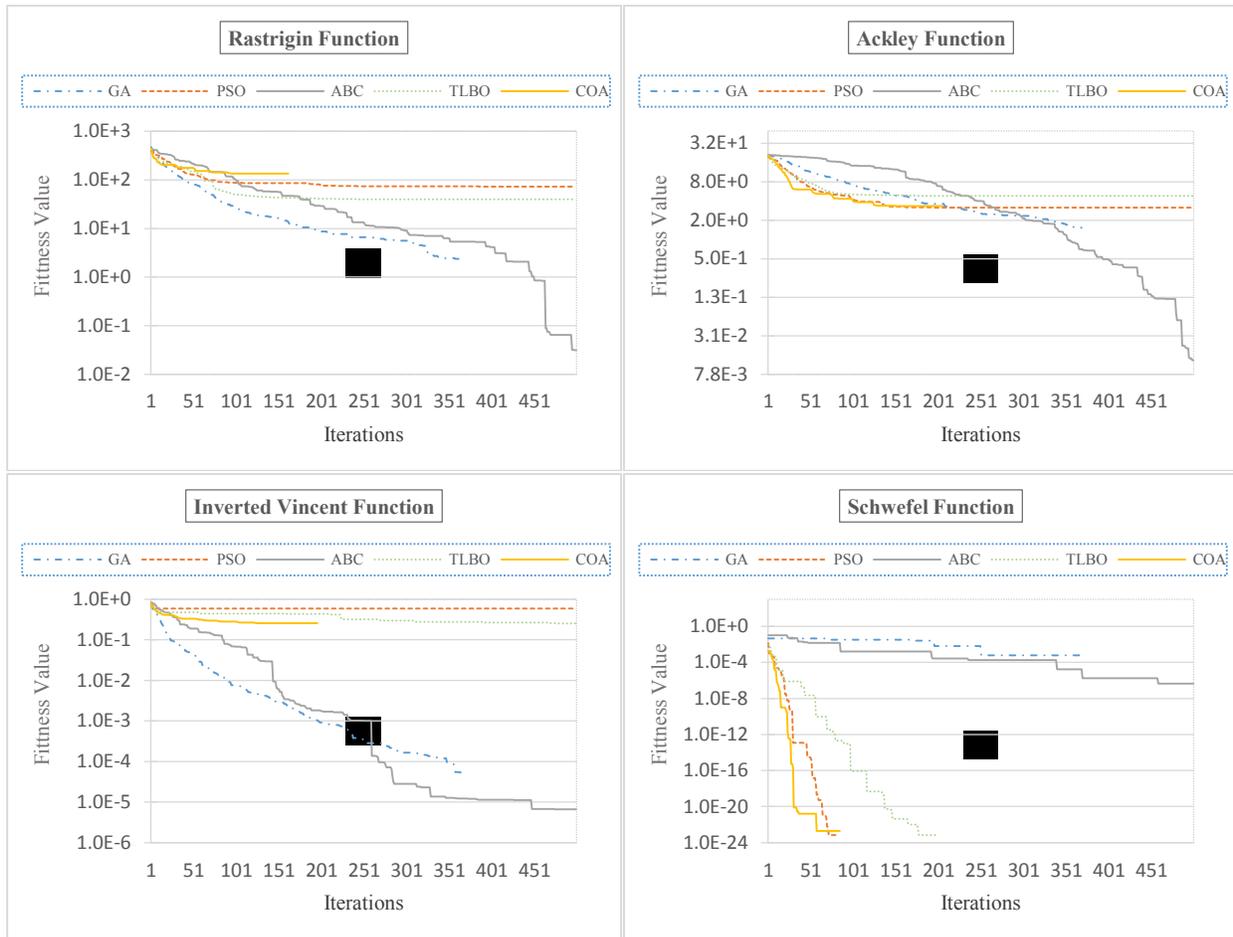

Fig. 5: Convergence diagram of the metaheuristics for four multimodal functions with the dimension of 30 (Top-Left: Rastrigin function, Top-Right: Ackley function, Bottom-Left: Inverted Vincent function and Bottom-Right: Schwefel function).

## 5. CONCLUSION

In this paper, a comprehensive empirical study on some basic metaheuristics i.e. Genetic Algorithm (GA), Particle Swarm Optimization (PSO), Artificial Bee Colony (ABC), Teaching-Learning-Based Optimization (TLBO), and Cuckoo Optimization algorithm (COA) was conducted, and various results and conclusions were obtained. The achieved results were analyzed from two different points of view; On one hand, from the performance perspective, the results revealed to us the following conclusions. Firstly, TLBO had a significant performance on the minimization benchmark functions of which their global optimum is located on $[0, 0, …,0]^T$, while when the global minimum was relocated randomly, it lost much of its efficiency, and this was as a consequence of a bias movement toward the center of search space in TLBO. Secondly, ABC is strongly suggested for complex real-world combinatorial optimization problems since it showed an outstanding performance in both unimodal and multimodal benchmark functions. Thirdly, although PSO was very good on unimodal functions, it gave up its $2^{nd}$ rank to GA on multimodal benchmark functions; however, PSO was very faster, but GA consume a huge time and resource to be performed so that we do not suggest it where time is not convenient for the application e.g. real-time decision or communication. Finally, considering all the experiments on shift-rotated of both unimodal and multimodal benchmark functions leaded us to the following ranking, {ABC, PSO, TLBO, GA,

COA}. On the other hand, from the convergence perspective, plotted convergence graphs for both unimodal and multimodal functions revealed to us the following conclusions. First of all, GA and COA have terminated before a complete 40000 FFEs in some cases, while others not, and this justifies their occasional degraded results versus PSO, ABC, and TLBO; on this basis, we suggest that they are adequate for those problems needing fast decisions e.g. real-time applications or communications; nevertheless, some extra mutation mechanisms may have a good contribution in their performance, too. Secondly, among PSO, ABC, and TLBO, which exploit all of their accessible FFEs, PSO is very good in unimodal functions while ABC has significant convergences in multimodal ones; hence, we strongly suggest the utilization of PSO in simple engineering problems, where the problem can be modeled as a standard unimodal optimization problem, while applying the ABC on the more complex engineering and industrial problems, where the search space has some global optima among a number of local minima/maxima around.

## REFERENCES

Adam P. Piotrowski, Maciej J. Napiorkowski, Jaroslaw J. Napiorkowski, Pawel M. Rowinski, "Swarm Intelligence and Evolutionary Algorithms: Performance versus speed," Information Sciences 384 (2017), 34-85.

Ahrari, Ali, and Ali A. Atai. "Grenade explosion method—a novel tool for optimization of multimodal functions." Applied Soft Computing 10, no. 4 (2010): 1132-1140.

Akbari, Mehdi, and Hassan Rashidi. "A multi-objectives scheduling algorithm based on cuckoo optimization for task allocation problem at compile time in heterogeneous systems." Expert Systems with Applications 60 (2016): 234-248.

Bazgosha, Atiyeh, Mohammad Ranjbar, and Negin Jamili. "Scheduling of loading and unloading operations in a multi stations transshipment terminal with release date and inventory constraints." Computers & Industrial Engineering 106 (2017): 20-31.

Dorigo, Marco, Vittorio Maniezzo, Alberto Colorni, and Vittorio Maniezzo. "Positive feedback as a search strategy." (1991).

Elyasigomari, V., D. A. Lee, H. R. C. Screen, and M. H. Shaheed. "Development of a two-stage gene selection method that incorporates a novel hybrid approach using the cuckoo optimization algorithm and harmony search for cancer classification." Journal of Biomedical Informatics 67 (2017): 11-20.

Eusuff, Muzaffar M., and Kevin E. Lansey. "Optimization of water distribution network design using the shuffled frog leaping algorithm." Journal of Water Resources planning and management 129, no. 3 (2003): 210-225.

Farmer, J. Doyne, Norman H. Packard, and Alan S. Perelson. "The immune system, adaptation, and machine learning." Physica D: Nonlinear Phenomena 22, no. 1-3 (1986): 187-204.

Faradonbeh, Roohollah Shirani, and Masoud Monjezi. "Prediction and minimization of blast-induced ground vibration using two robust meta-heuristic algorithms." Engineering with Computers (2017): 1-17.

Fogel, L.J., Owens, A.J. and Walsh, M.J., Artificial Intelligence Through Simulated Evolution, John Wiley, New York, USA (1966).

Geem, Zong Woo, Joong Hoon Kim, and G. V. Loganathan. "A new heuristic optimization algorithm: harmony search." Simulation 76, no. 2 (2001): 60-68.

Holland, J. H. Adaptation in Nature and Artificial Systems. The University of Michigan press, USA, (1975).

IEEE Congress on Evolutionary Computation 2017 (IEEE CEC'2017), http://www.cec2017.org, Donostia - San Sebastián, Spain, June 5-8, 2017.

Karaboga, Dervis. An idea based on honey bee swarm for numerical optimization. Vol. 200. Technical report-tr06, Erciyes university, engineering faculty, computer engineering department, 2005.

Kashan, Ali Husseinzadeh. "An efficient algorithm for constrained global optimization and application to mechanical engineering design: League championship algorithm (LCA)." Computer-Aided Design 43, no. 12 (2011): 1769-1792.

Kaveh, A., and S. Talatahari. "A novel heuristic optimization method: charged system search." Acta Mechanica 213, no. 3 (2010): 267-289.

Kennedy, J., and R. Eberhart. "Particle swarm optimization (PSO)." In Proc. IEEE International Conference on Neural Networks, Perth, Australia, pp. 1942-1948. 1995.

Li, Xiaodong, Ke Tang, Mohammad N. Omidvar, Zhenyu Yang, Kai Qin, and Hefei China. "Benchmark functions for the CEC'2013 special session and competition on large-scale global optimization." gene 7, no. 33 (2013): 1-8.

Liang, Jane-Jing, Ponnuthurai Nagaratnam Suganthan, and Kalyanmoy Deb. "Novel composition test functions for numerical global optimization." In Swarm Intelligence Symposium, 2005. SIS 2005. Proceedings 2005 IEEE, pp. 68-75. IEEE, 2005.

Liang, J. J., B. Y. Qu, P. N. Suganthan, and Alfredo G. Hernández-Díaz. "Problem definitions and evaluation criteria for the CEC 2013 special session on real-parameter optimization." *Computational Intelligence Laboratory, Zhengzhou University, Zhengzhou, China and Nanyang Technological University, Singapore, Technical Report* 201212 (2013): 3-18.

Awad, N. H., M. Z. Ali, J. J. Liang, B. Y. Qu, and P. N. Suganthan. "Problem definitions and evaluation criteria for the CEC 2017 special session and competition on single objective bound constrained real-parameter numerical optimization." In *Technical Report*. NTU, Singapore, 2016.

Nabaei, Armin, Melika Hamian, Mohammad Reza Parsaei, Reza Safdari, Taha Samad-Soltani, Houman Zarrabi, and A. Ghassemi. "Topologies and performance of intelligent algorithms: a comprehensive review." *Artificial Intelligence Review* (2016): 1-25.

Passino, Kevin M. "Biomimicry of bacterial foraging for distributed optimization and control." IEEE control systems 22, no. 3 (2002): 52-67.

Rajabioun, Ramin. "Cuckoo optimization algorithm." Applied soft computing 11, no. 8 (2011): 5508-5518.

Rao, Ravipudi V., Vimal J. Savsani, and D. P. Vakharia. "Teaching–learning-based optimization: a novel method for constrained mechanical design optimization problems." Computer-Aided Design 43, no. 3 (2011): 303-315.

I. Rechenberg, Evolutionsstrategie: Optimierung Technischer Systeme nach Prinzipien der Biologischen Evolution, Frommann-Holzboog, Stuttgart, Germany, 1973.

Rashedi, Esmat, Hossein Nezamabadi-Pour, and Saeid Saryazdi. "GSA: a gravitational search algorithm." Information sciences 179, no. 13 (2009): 2232-2248.

Simon, Dan. "Biogeography-based optimization." IEEE transactions on evolutionary computation 12, no. 6 (2008): 702-713.

Storn, Rainer, and Kenneth Price. "Differential evolution–a simple and efficient heuristic for global optimization over continuous spaces." Journal of global optimization 11, no. 4 (1997): 341-359.

Wilcoxon, Frank. "Individual comparisons by ranking methods." Biometrics bulletin 1, no. 6 (1945): 80-83.

Wolpert, David H., and William G. Macready. "No free lunch theorems for optimization." *IEEE transactions on evolutionary computation* 1, no. 1 (1997): 67-82.